\documentclass{article}

\usepackage{times}
\usepackage{graphicx} % more modern
\usepackage{subfigure} 

\usepackage{natbib}

\usepackage{algorithm}
\usepackage{algorithmic}

\usepackage{enumerate}

\usepackage{hyperref}

\usepackage[accepted]{icml2017}

\icmltitlerunning{SenGen: Sentence Generating Neural Variational Topic Model}

\begin{document} 

\twocolumn[
\icmltitle{SenGen: Sentence Generating Neural Variational Topic Model}

% It is OKAY to include author information, even for blind
% submissions: the style file will automatically remove it for you
% unless you've provided the [accepted] option to the icml2017
% package.

% list of affiliations. the first argument should be a (short)
% identifier you will use later to specify author affiliations
% Academic affiliations should list Department, University, City, Region, Country
% Industry affiliations should list Company, City, Region, Country

% you can specify symbols, otherwise they are numbered in order
% ideally, you should not use this facility. affiliations will be numbered
% in order of appearance and this is the preferred way.
\icmlsetsymbol{equal}{*}
%\icmlsetsymbol{doneat}{+}
\begin{icmlauthorlist}
\icmlauthor{Ramesh Nallapati}{amazon}
\icmlauthor{Igor Melnyk}{ibm}
\icmlauthor{Abhishek Kumar}{ibm}
\icmlauthor{Bowen Zhou}{ibm}
\end{icmlauthorlist}

\icmlaffiliation{ibm}{AI Foundations, IBM Research}
\icmlaffiliation{amazon}{AWS Deep Learning, Amazon (Entire Work done at IBM Research)}
%\icmlaffiliation{doneat}{Entire work done while at IBM Research,}

\icmlcorrespondingauthor{Ramesh Nallapati}{rnallapa@amazon.com}
%\icmlcorrespondingauthor{Eee Pppp}{ep@eden.co.uk}

% You may provide any keywords that you 
% find helpful for describing your paper; these are used to populate 
% the "keywords" metadata in the PDF but will not be shown in the document
\icmlkeywords{VAE, Topic Modeling, Document generation}

\vskip 0.3in
]

% this must go after the closing bracket ] following \twocolumn[ ...

% This command actually creates the footnote in the first column
% listing the affiliations and the copyright notice.
% The command takes one argument, which is text to display at the start of the footnote.
% The \icmlEqualContribution command is standard text for equal contribution.
% Remove it (just {}) if you do not need this facility.

\printAffiliationsAndNotice{}  % leave blank if no need to mention equal contribution
%\printAffiliationsAndNotice{\icmlEqualContribution} % otherwise use the standard text.

\begin{abstract}
We present a new topic model that generates documents by sampling a topic for one whole sentence at a time, and generating the words in the sentence using an RNN decoder that is conditioned on the topic of the sentence. We argue that this novel formalism will help us not only visualize and model the topical discourse structure in a document better, but also potentially lead to more interpretable topics since we can now illustrate topics by sampling representative sentences instead of bag of words or phrases. We present a variational auto-encoder approach for learning in which we use a factorized variational encoder that independently models the posterior over topical mixture vectors of documents using a feed-forward network, and the posterior over topic assignments to sentences using an RNN. Our preliminary experiments on two different datasets indicate early promise, but also expose many challenges that remain to be addressed.
\end{abstract}

% The \author macro works with any number of authors. There are two commands
% used to separate the names and addresses of multiple authors: \And and \AND.
%
% Using \And between authors leaves it to \LaTeX{} to determine where to break
% the lines. Using \AND forces a linebreak at that point. So, if \LaTeX{}
% puts 3 of 4 authors names on the first line, and the last on the second
% line, try using \AND instead of \And before the third author name.

%\iclrfinalcopy % Uncomment for camera-ready version

%\maketitle

\section{Introduction}
One of the most popular approaches for fully generative modeling of documents is the Latent Dirichlet Allocation \cite{lda} model. This model that assumes a discrete mixture distribution over topics for each document that is sampled from a Dirichlet prior shared by all documents. A topic is sampled for each word position in the document from this mixture and then the word itself is generated from another multinomial indexed by the corresponding topic. Although very successful in various tasks, one of the shortcomings of this model is its bag-of-words approach where dependencies between words are not explicitly modeled. 
%not capture short term and long-term capture dependencies between words in a document explicitly, except through the shared topics. 
Several extensions of LDA have been proposed to relax the bag-of-words assumption and capture longer term relationships between words \cite{hmm_lda,topical_ngrams,seq_lda}, segments \cite{unsup_topic_seg}, and discourse elements in documents \cite{bayesian_sent_topic_discourse,conversation_trees}. 

%With the advent of neural networks, RNN-based language models have emerged as the de-facto models to capture short and long range dependencies between words \cite{rnnlm}. However, these models do not capture the high-level topical context or the discourse structure of the document.

Recently, Kingma and Welling proposed a Variational Auto-Encoder (VAE) based approach for learning complex generative distributions where the generative model as well as the approximate variational posterior are based on deep neural networks \cite{vae}. This approach has been recently applied to topic modeling of documents by several researchers. One of the first VAE-based approaches for document modeling is called the Neural Variational Document Model (NVDM) \cite{nvdm}, which reports impressive gains over LDA and other models on perplexity. 
%Drawing analogy to the LDA model, the equivalent of the topic mixture for a document in NVDM is ${\bf h}$, a K-dimensional real-valued vector drawn from a multi-variate Gaussian representing the `strength' of various topics in the document. Each word is then drawn from a softmax layer whose unnormalized score is based on the dot product between ${\bf h}$ and the word's embedding of size $K$. In LDA parlance, the dot product is equivalent to marginalizing the topic assignment variable $z_i$ for each word position $i$ in the document. The Variational posterior for ${\bf h}$ is also a Gaussian whose mean and variance are obtained from a feed-forward neural network that accepts the counts vector of all the words in the document as input. The reason Gaussian prior and posterior are chosen is that it allows a closed form computation of the KL-divergence term in the variational lower bound, besides permitting the Reparameterization Trick (RT) \cite{vae} for the reconstruction error term in the lower bound. 
However the topic mixture vector in this model, ${\bf h}$, being a real-valued  vector generated from a multivariate Gaussian, is not very interpretable unlike multinomial mixture in the standard LDA model. Motivated by this weakness of NVDM, the authors of \cite{avi_tm} propose the NVLDA model that employs a Logistic Normal distribution to replace the Dirichlet prior and a variational Logistic Normal posterior to bring the ${\bf h}$ vector into the multinomial space. 
%This formulation allows closed form computation for the KL-divergence term, as well as application of RT in the reconstruction error term of the variational lower bound. 
However, the perplexity values from the new NVLDA model on unseen data are worse than those from the NVDM model. Although both the models mentioned above employ sophisticated VAE approach, they still use the same bag-of-words formalism of LDA in modeling the document. Further, their VAE approach focuses only on modeling the posteriors over the document-level topic mixtures vector, and ignores modeling the posteriors over the local topic assignment to words and sentences.

With the advent of neural networks, RNN-based language models have emerged as {\it de facto} choice to capture short and long range dependencies between words \cite{rnnlm} and have been used for language modeling in speech \cite{vrnn}, and dialogue \cite{hrnn_dialogue}. However, these models do not capture the topical structure of the larger document context. A recent work that integrates topic modeling with RNNs is that of \cite{topic_rnn}, where a Gaussian based topic vector, similar to the one used  in NVDM, is used to model topic strengths for each document, but an RNN is used to generate words conditioned on the topic vector. 
%The variational posterior for the topic strengths vector used is also similar to NVDM, and the expected reconstruction error term from the RNN in the variational lower bound is computed using the reparameterization trick on a Gaussian posterior. 
The Topic RNN model marginalizes the topic assignments to words, without explicitly modeling their posteriors.

\section{SenGen: Sentence Generating Topic Model}
  
Similar to the work of \cite{topic_rnn}, we are interested in modeling dependencies between words in a document and also capturing the larger topical context jointly.  In addition, we are also interested in capturing the topical discourse structure in a document including notions such as topical drift and topical switch. To capture such phenomena, we argue that sentences are the ideal smallest units for modeling instead of individual words or phrases, since sentences tend to be topically cohesive while topical drift or switch usually occur across sentence or paragraph boundaries. We therefore make topical assignments to whole sentences, unlike traditional topic models that assign them to each individual word position in the document. We use RNNs to generate the words in each sentence conditioned on its assigned topic, so as to capture within-sentence dependencies between words. We believe our modeling choice not only allows us to better visualize topical discourse structure in a document (say, by analyzing the changes in the posterior over the topic assignment variables for sentences as move from start to end of the document), but may also potentially lend topics better interpretability since we can visualize them by generating representative sentences from the learned topic-specific RNN word generators.  

 In this work, we will also present a VAE framework to model the posteriors over the topic assignment variables at sentence-level explicitly through an encoder based on another RNN. Previous work on VAE-based learning approach for topic models such as \cite{topic_rnn}, \cite{nvdm} and \cite{avi_tm} focus on modeling the posterior of the topical mixture at document-level, but ignore the issue of modeling the posterior of topic assignments. We hope that our work on explicit modeling of the posteriors of the topic assignment variables will pave the way for future work on more sophisticated posteriors that can also capture topical correlations across neighboring sentences.
 
 %To model the topic strengths vector at document level, we retain the framework of \cite{topic_rnn} and \cite{nvdm}. %We propose several novel methods to model the posterior including explicit and implicit distributions.

\subsection{Generative Process}
The {\it SenGen} model first samples the document-level topic strengths $\theta_d$ from a $K$-dimensional multivariate Gaussian $\mathcal{N}(\cdot|{\bf 0},{\bf I})$. Topic indices $z_s$ are sampled from the mixture distribution $\texttt{softmax}(\theta_d)$ for {\it each sentence} $s$ in the document. Conditioned on the topic-id $z_s$, a topic-specific GRU-RNN \cite{gru} based decoder is run to generate all the words ${\bf w}_s$ in the sentence. The conditioning on topic is done via a topic-embedding vector \texttt{EmbZ} which is also learned automatically. 

In effect, our model relaxes the bag-of-words assumption of LDA, and instead assumes the document to be a bag of independent sentences, each of which can assume its own topic. The words in each sentence share the same topic, and are generated jointly using the decoder RNN. %As mentioned earlier, one of the advantages of this approach is that we can now represent each topic by generating  sample sentences from that topic, which may lend more interpretability than the bag-of-words representations. Further, our inference techniques that estimate the posterior over topics for {\it each sentence} allow us to visualize the topical discourse of the document at the sentence-level, which we believe is at the right level of granularity to study phenomena such as topic drift or topic switch.

\begin{figure}[ht]
    \vspace{-0.1in}
	\centering
  \includegraphics[width=\columnwidth]{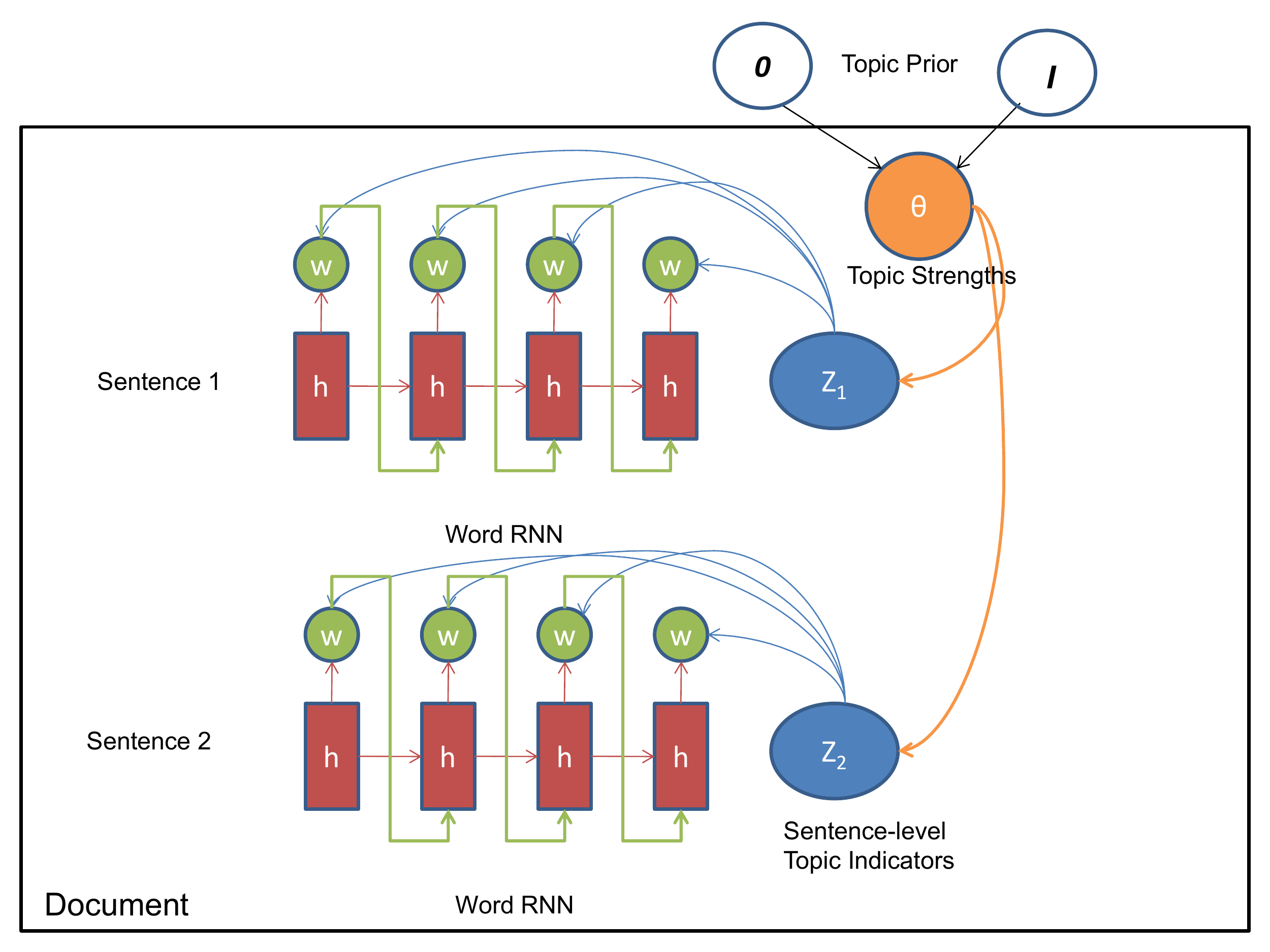}
  	\vspace{-0.3in}
	\caption{{\small Graphical representation of the {\it SenGen} Model. }}
	\label{fig:topic_rnn}
\end{figure}

The steps in Algorithm \ref{tab:gen_process} describe the generative process of the model in more detail. A graphical representation of the model is also presented in Figure \ref{fig:topic_rnn}. Note that there is a separate decoder RNN for each topic, but they all share the same parameters except for the word-generating softmax layer. %The word embeddings $\texttt{Emb}$, topic embeddings $\texttt{EmbZ}$ are the additional parameters of the model which are learned automatically at training time. 

\begin{algorithm*}[tb]
\caption{Generative process for {\it SenGen} Model}
\label{tab:gen_process}
%\vspace{-0.3in}
%\begin{algorithmic}
\begin{enumerate}
\item For each document $d$ in $\{1,\cdots,N\}$:
\item \hspace{0.2in} Generate un-normalized topic mixture of the document $\theta_d \sim \mathcal{N}(\cdot|{\bf 0}, {\bf I})$
\item \hspace{0.2in} For each sentence $s$ in $\{1,\cdots,N_s^{(d)}\}$:
\item \hspace{0.4in} Sample topic $z_s \sim \texttt{Mult}(\texttt{softmax}(\theta_d))$
\item \hspace{0.4in} Initialize the hidden state of the RNN as ${\bf h}_0^{s} = \texttt{zeros}(|{\bf h}|)$
\item \hspace{0.4in} Set the embeddings of the zeroth word in the RNN as $\texttt{Emb}[w_{0}] = \texttt{zeros}(|\texttt{Emb}|)$
\item \hspace{0.4in} Select context vector for the RNN from topic embeddings: ${\bf c}_s = \texttt{EmbZ}[z_s]$
\item \hspace{0.4in} For each word position in the sentence $i$ in $\{1,\cdots,N_w^{(s)}\}$:
\item \hspace{0.6in} Update the hidden state of $\texttt{RNN}$ ${\bf h}_i = \texttt{tanh}(W_h{\bf h}_{i-1}+W_e\texttt{Emb}[w_{i-1}]+W_c {\bf c}_s+b)$
\item \hspace{0.6in} Compute the readout layer ${\bf r}_i = \texttt{tanh}(W^r_{h}{\bf h}_i  + W^r_e\texttt{Emb}[w_{i-1} ] + W^r_{c}{\bf c}_s + b^r)$
\item \hspace{0.6in} Generate word using $P(w_i=w^{v}) = \texttt{softmax}(W_{z_s}^v{\bf r}_i + b^v)$
\end{enumerate}
%\end{algorithmic}
\end{algorithm*}

% $\texttt{Emb}[w_i]$ is the embedding of word $w_i$, which is automatically learned by the model, and $W_{h,\cdot}$ and $W_{c,\cdot}$ are the parameters of the \texttt{softmax} layer, while ${\bf h}_i$ and ${\bf c}_s$ represent the hidden states of word-level RNN and sentence-level RNN respectively. The term $W_{z_s}^T\texttt{Emb}(w_i)$ captures the importance of term $w_i$ in topic $z_s$.

\subsection{Likelihood and Parameter Learning}

The observed data log likelihood of a document corpus $C=\{d_1,\cdots,d_N\}$ from this model is given by:
\begin{eqnarray}\label{eq:lhood}
&&P({\bf w}|\beta) = \prod_{d=1}^N\int_{\theta_d}\mathcal{N}(\theta_d|{\bf 0} ,{\bf I}) ( \nonumber\\
&&\prod_{s=1}^{N_s^{(d)}}\sum_{z_s=1}^K P(z_s|\theta_d)P({\bf w}_s|z_s,\beta))d\theta_d
\end{eqnarray}
where
\begin{equation}
P(z=k|\theta_d) = \texttt{softmax}(\theta_{dk}),
\end{equation}
and $\beta$ are the parameters of the word-generating RNN decoder and $\texttt{softmax}(x_k) = \frac{\exp(x_k)}{\sum_{k'}\exp(x_{k'})}$ is an operator that maps the topic strengths vector $\theta_d$ into a multinomial simplex, and $K$ is a hyperparameter indicating the number of topics in the model, and $N_s^{(d)}$ is the number of sentences in document $d$.

The likelihood of the words ${\bf w}_s$ in each sentence using the RNN is given by:
\begin{eqnarray}
P({\bf w}_s|z_s,\beta) &=& \prod_{i=1}^{N_w^{(s)}}P(w_i|{\bf w}_{-i},z_s,\beta),
%P(w_i|{\bf w}_{-i},z_s,\texttt{RNN}_s,\texttt{RNN}_w) &=& \texttt{softmax}(W_{h,w_i}^T{\bf h}_i+ W_{c,w_i}^T{\bf c_s} + W_{z_s}^T\texttt{Emb}(w_i))
\end{eqnarray}
where each term in the RHS of the equation above is computed using step (11) in the generative process displayed in Algorithm \ref{tab:gen_process}. 

\subsection{Learning using VAE}

We consider the following factorized variational encoders to model the posteriors for the latent variables $\theta$ for documents and $z$ for sentences. 
\begin{eqnarray}
q(\theta_1,{\bf z}_1,\cdots,\theta_N,{\bf z}_N|{\bf w}) &=&\prod_{d=1}^N q(\theta_d|{\bf w}_d)\prod_{s=1}^{N_s^{(d)}}q(z_s|{\bf w}_s)\nonumber\\
\end{eqnarray}
where ${\bf w}_d$ is the vector of words in the document $d$ and ${\bf w}_s$ is the vector of words in the sentence $s$. In other words, the posteriors over the topic indicators for each sentence are assumed to be independent from the posteriors of the topic vector for the entire document. This is clearly a simplifying assumption that makes inference tractable, and may need to be relaxed in the future.

Note that the encoders are amortized over all the documents unlike mean-field approaches where each latent variable is assumed to have its own independent posterior \cite{lda}.
 The document-level encoder $q(\theta_d|{\bf w}_d)$ is a simple feed forward network that estimates the mean and co-variance of the posterior for $\theta_d$ as given by the following series of steps:
\begin{eqnarray}
\gamma_d &=& \texttt{tanh}(W_{\gamma}(\sum_{i=1}^{N^{(d)}_w}\texttt{Emb}[w_i]) + b_{\gamma}) \nonumber\\
\mu_d &=& W_{\mu}\gamma_d+b_{\mu} \nonumber\\
{\bf \sigma}_d &=& \texttt{exp}(W_{\sigma}\gamma_d+b_{\sigma}) \nonumber\\
\hat{\theta}_d &=& \mu_d + {\bf \sigma}_d\odot\epsilon
\end{eqnarray}
where $N^{(d)}_w$ is the number of words in the document and $W_{\gamma}, b_{\gamma}, W_{\mu}, b_{\mu}, W_{\sigma}, b_{\sigma}$ are the parameters of the encoder. $\epsilon$ is a $K$-dimensional Gaussian noise vector generated from $\mathcal{N}(0,1)$. In the last equation above, we used the reparametrization trick to sample $\hat{\theta}_d$ from the encoder's posterior $q(\theta_d|{\bf w}_d)$ while maintaining end-to-end differentiability of the model.

\begin{figure}[ht]
    \vspace{-0.1in}
	\centering
  \includegraphics[width=\columnwidth]{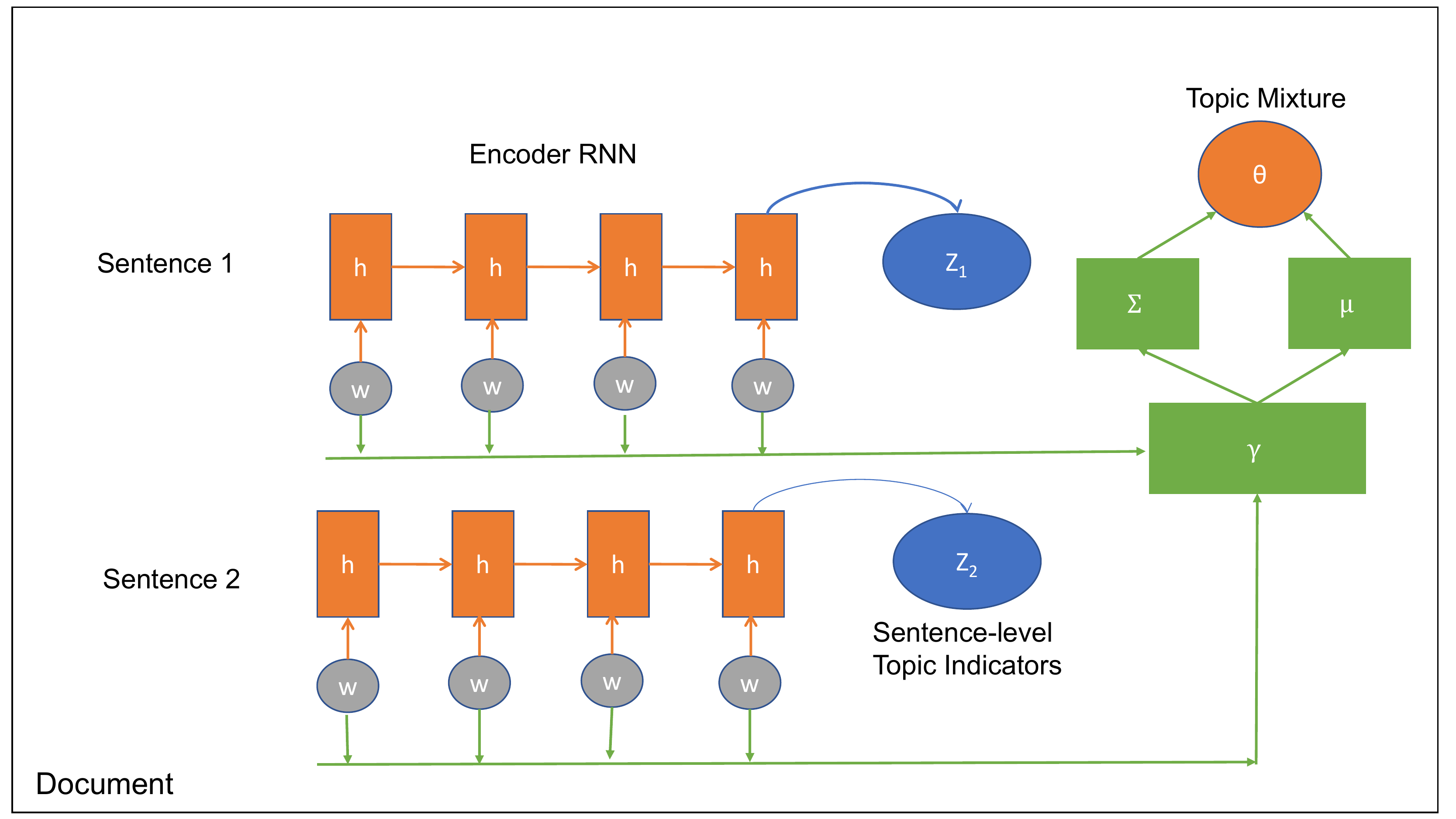}
  	\vspace{-0.3in}
	\caption{{\small Graphical representation of the encoder architecture: the posteriors for sentence-level topic assignments $z_s$ and for document level topical strengths $\theta_d$ are modeled as independent of each other.}}
	\label{fig:variational_encoder}
\end{figure}

The sentence-level encoder is another GRU-RNN that outputs the posterior over topics given the words in the sentence ${\bf w}_s$ as follows.
\begin{eqnarray}
 &&q(z_s=k|{\bf w}_s) = \texttt{softmax}(W^{(enc)}_{k}\cdot{\bf h}^{(enc)}_{N_w^{(s)}}+b_k),\nonumber\\
 &&\texttt{where}~ {\bf h}^{enc}_{i} = \texttt{GRU}({\bf h}^{(enc)}_{i-1},\texttt{Emb}[w_i])
\end{eqnarray}
Thus, the encoder RNN consumes all the words in the sentence as input, one at every time step, and emits the posterior probabilities over topics at the last time step $N^{(s)}_w$. The graphical representation of both the variational encoders is displayed in Figure \ref{fig:variational_encoder}.

Given the two encoders, the variational lowerbound for the log-likelihood of the observed data in the VAE approach can be written as:
\begin{eqnarray}\label{eq:elbo}
\log P({\bf w}_d|\beta) &\ge& -KL(q(\theta_d)\|P(\theta_d)) + \nonumber\\
&& \sum_{s=1}^{N_s^{(d)}}(E_q\log P(z_s|\theta_d) +H(q(z_s|{\bf w}_s)) \nonumber\\
&&  + E_q\log P({\bf w}_s|z_s,\beta))
\end{eqnarray}
Further, each term on the RHS of Eq. (\ref{eq:elbo}) can be factorized as follows.
\begin{eqnarray}\label{eq:elbo_terms}
KL(q(\theta_d)\|P(\theta_d)) &=& \frac{1}{2}\sum_{k=1}^K(1+\log((\sigma)_{j}^2)-\mu_j^2-\sigma_{j}^2)\nonumber\\
E_q\log P(z_s|\theta_d) &\approx& \sum_k \log \frac{\exp(\hat{\theta}_{dk})}{\sum_{k'}\exp(\hat{\theta}_{dk'})} \nonumber \\
H(q(z_s|{\bf w}_s)) &=& \sum_k -q(k|{\bf w}_s)\log q(k|{\bf w}_s)\nonumber\\
E_q\log P({\bf w}_s|z_s,\beta) &=& \sum_k q(k|{\bf w}_s)\log P({\bf w}_s|\beta,k) 
\end{eqnarray}
where the first term above, involving KL-divergence between two Gaussians, is computed analytically as described in \cite{vae}. The second term above, consisting of the expectation of $P(z_s|\theta_d)$ is computed using a sample based estimate.
For computing the other terms, we took advantage of the clear separation of ${\bf z}_s$ and $\theta_d$ in the posterior to compute the expectations exactly, and involves summation over all topics. 
\section{Experiments}

\subsection{Datasets}
In our experiments, we used the `by-date' version of the 20 Newsgroups dataset downloadable from \url{http://qwone.com/~jason/20Newsgroups/} as well as the CNN/Daily Mail corpus available at \url{http://cs.nyu.edu/~kcho/DMQA/}. 

Although preprocessed versions of the 20 Newsgroups datasets are available where the text is tokenized and words converted to integer ids, we preprocessed the text on our own since we needed to preserve sentence boundary information. We did not remove any stopwords since we want the model to produce meaningful sentences. We used the official training and test splits defined in the {\it by-date} version of the dataset. We further sub-divided training set into  training and validation sets, so that the validation loss could be used for early stopping of the training process. In all, we had 10,000 training documents, 1,314 validation documents and 7,532 test documents. There are about 18 sentences per document and 20 words per sentence on average. We pruned our vocabulary to 60,359 most frequent words, which is very close to that reported in other experiments \cite{avi_tm}. However, it is not clear to us if the vocabulary we used in our experiments is identical to the vocabulary in the preprocessed versions, since we had to do many clean-up operations such as removing email-headers and signatures from the documents to reduce noise.

The CNN/Daily Mail corpus is a large corpus consisting of more than 300,000 documents. This documents in this corpus are well formatted with sentence boundaries, which is required in our model. We randomly subsampled 10,000 documents for training, 1,000 documents for validation and 2,000 documents for testing. On average, this dataset has 29 sentences per document, and 26 words per sentence. The vocabulary size is pruned to 55,226 top most frequent words. We ran the other baselines on this corpus using their open source code. 

The authors of \cite{nvdm} and \cite{avi_tm} use the RCV1 corpus as an additional corpus, but the free version of this corpus\footnote{\url{http://www.ai.mit.edu/projects/jmlr/papers/volume5/lewis04a/lyrl2004_rcv1v2_README.htm}} is already tokenized with no sentence boundary information, hence we ignored it in our experiments.

\subsection{Model settings}
For runs on both datasets, we used word embeddings of dimension 100, pre-trained using {\it word2vec} on the full CNN/Daily Mail corpus. We set the hidden state of both the encoder and decoder RNNs to 200, the dimension of the readout layer to 100. Each topic has its own decoder but they all share the same parameters except in the softmax layer where the parameters are distinct. Since the softmax layer is of size $[ |V| \times |r| ]$ where $ |V| $ is the training vocabulary size, and $|r|$ is the size of the readout layer (see steps 10 and 11 in Table \ref{tab:gen_process}), training the model is very challenging both in terms of space and time computational requirements. Therefore, we limited our number of topics to 25, and also fixed our batch size to 1. 

\begin{table*}[ht]
\begin{center}
\begin{tabular}{|l|r|r|}
\hline
Model & 20 Newsgroups & CNN/Daily Mail \\
\hline
LDA \cite{lda} & 1247 & 776\\
NVDM \cite{nvdm} & 757 & 435 \\
NVLDA \cite{avi_tm} & 1213 & 592\\
ProdLDA \cite{avi_tm} & 1695 & 735 \\
%Topic-RNN & N/A & N/A \\
SenGen (Our Model) & 2354 & 671 \\
\hline
\end{tabular}
\caption{Perplexity comparison of various models on two different datasets. All models are configured to use 25 topics. Lower is better.}
\label{tab:perplexity}
\end{center}
\end{table*}

To save GPU memory, we implemented a variant of the large vocabulary trick \cite{lvt} where for each batch we sampled a subset of 4,000 words from the training corpus distribution to be used as the vocabulary in the softmax layer, in addition to the words that occurred in that batch. Despite this, training is still very slow since we need to compute the softmax probabilities for each sentence from all $K$ decoders. On the 20 Newsgroups dataset, the total number of parameters of our model were 159M, and each epoch took 6.5 hours on an average on a single K80 GPU. On the CNN/Daily Mail corpus, our parameter size is 145M and each epoch took 8.3 hours. Although CNN/Daily Mail corpus has smaller number of parameters owing to its smaller vocabulary size, it has longer documents on average than 20 Newsgroups, which explains the longer training time. We used {\it Adadelta} variant of the SGD that automatically adapts the learning rate, and we used gradient clipping for training stability. We did not use any regularization in the objective function. %We trained our model for only 5 epochs on both datasets due to slow training time.

To reduce the model size, we also experimented with a variant of our model where even the softmax layer is shared by all topic-specific decoders except a bias vector which is topic-specific. This new model resulted in considerable memory savings and also sped up training time due to smaller number of parameters. However it failed to learn distinct topics which forced us to abandon this variant.

\subsection{Perplexity results}

 We compute perplexity of the test dataset using the trained {\it SenGen} model as follows:
\begin{equation}
    \texttt{Perplexity} = \frac{1}{N}\sum_{d=1}^{N}\exp(\frac{-\log P({\bf w}_d|\beta)}{N_d})
\end{equation}
where the log probability is computed using the lower-bound estimate in Eq. (\ref{eq:elbo}). In the above equation, $N$ is the number of test documents, and $N_d$ is the number of words in document $d$.  We compared the perplexity of our model with models from \cite{lda}, \cite{nvdm} and \cite{avi_tm}. We could not compare our results with \cite{topic_rnn} since they reported their numbers on different datasets, and their code is not yet publicly released either. The results in Table \ref{tab:perplexity} indicate that the new {\it SenGen} model does not achieve better perplexity than any of the models we compared with on the 20 Newsgroups dataset. On CNN/Daily Mail, our model achieves better perplexity than LDA as well as ProdLDA variant of \cite{avi_tm}, but is not as good as the other VAE-based models. On 20 Newsgroups datasets, we suspect the main reason is due to the differences in preprocessing between our work and that of others -- we noticed that there are many non-dictionary terms in our vocabulary that originated from email signatures and headers. Another potential reason is that the model may be overfitting the training set due to the extremely large number of parameters. %In the future, we intend to investigate various forms of regularization to reduce overfitting.

\subsection{Qualitative results}

One advantage of assigning topics to whole sentences is that the decoder RNN learns to generate sentences for each topic, which could be potentially more interpretable than representing topics merely by top ranking words. In Table \ref{tab:qual}, we displayed the best sequences generated by beam search of width 5 on the decoder's softmax layer for three randomly chosen topics on the CNN / Daily Mail data set. We also displayed two different stochastic samples for the same topics where we greedily sample words from the distribution defined by the softmax layer of the RNN decoder for each time-step.

\begin{table*}
\begin{tabular}{|l|l|}
\hline
\multicolumn{2}{|c|}{{\bf Topic 1}} \\
\hline
Best sequence & he said that he had not been made with the case and that he had not been made with the case ... \\
Stochastic sample 1 &  bestsellers we positives hollywood-walk-of-fame dribbled in the association wilmington-10 horyn  ...\\
Stochastic sample 2 & timberlake united-lincolnshire-hospitals-trust enrolled snowballed helipad advertiser ...\\
\hline
\multicolumn{2}{|c|}{{\bf Topic 2}}\\
\hline
Best sequence & it is one of the world in the world of the world ... \\
Stochastic sample 1 & leader nasser chhattisgarh stroked arrogance debra-nelson impossibly fingers funding .. \\
Stochastic sample 2 & waterboarding pele will be compulsory to nh1 as department-of-defense darren-sammy scott-brown ... \\
\hline
\multicolumn{2}{|c|}{{\bf Topic 3}}\\
\hline
Best sequence & but it is not a lot of people who have to be able to be able to make ...\\
Stochastic sample 1 & catania ralph-lauren some0 impressionable re-interview texas-department-of-public-safety characters \\
Stochastic sample 2 & lucy-jones breakwater chats david-laws fanciful dyke gustafson said ... \\
\hline
\end{tabular}
\caption{Example sequences of words generated by our model trained on the CNN / Daily Mail corpus, conditioned on various topics. Best sequence is obtained by performing a beam search on the softmax layer of the decoder RNN. Stochastic samples are obtained by greedy sampling from the softmax layer of the RNN, one word at a time.}
\label{tab:qual}
\end{table*}

The table shows that the best sequences tend to be very generic, non-informative sentences. Although they are grammatically well formed in the beginning, they tend to repeat the generated phrases after a few time-steps. The stochastic samples, on the other hand, are not grammatically well formed, but do contain topical words. %Using our very biased perspective, we concluded that topic 1 is about celebrities, topic 2 is on politics and topic 3 discusses social interactions. 
	However the learned topics are certainly not as coherent as those learned by bag-of-words approaches such as LDA.

Clearly, more work needs to be done before we get these models learn more interpretable topics. To address the issue of non-informative best sequences, we may need to handle stop words and other frequent words in a special manner as done in the \cite{topic_rnn} work which used a separate class for these words which are then mixed with topics. Also, since the {\it SenGen} model has very large number of parameters, it may be desirable to initialize the model's parameters to those learned by a bag-of-words model, so that there is less chance it gets stuck at arbitrary local minima.

\section{Discussion and Conclusion}

The main contributions of this work are (i) assigning topics to whole sentences instead of words, so that the resulting topics have the potential to be more interpretable since we can generate representative sentences for each topic; (ii) presenting a VAE approach that not only models posteriors of topic mixtures at document-level but also of topic assignments at sentence-level.

Preliminary qualitative and quantitative results indicate some promise, but deeper investigation needs to be conducted to overcome some of the existing deficiencies of the current model such as handling frequent words, preventing overfitting, learning better topics and improving computational efficiency. 

Although one of our motivations is to capture topical discourse structure including the phenomena of topic drift and topic switch, this work addresses this issue only partially through the posteriors over topics for each sentence, which can be visualized graphically. We believe the framework proposed in this work can be extended to construct more sophisticated models that can capture dependencies between topics of adjacent sentences. Another direction we are interested in exploring is to provide the decoder with not only topical context but also the context from previous sentences in the document. Finally, we also need to relax the assumption of fully factorized posteriors of the document's topic vector and those of the sentence topics.

\pagebreak

\bibliographystyle{icml2017}
\bibliography{references}

\end{document}